\documentclass{article} 
\usepackage[preprint]{colm2025_conference}

\usepackage{microtype}
\usepackage{hyperref}
\usepackage{url}
\usepackage{booktabs}
\usepackage{tcolorbox}
\usepackage{xcolor}
\usepackage{soul}
\usepackage{lineno}
\usepackage{soul}

\definecolor{darkblue}{rgb}{0, 0, 0.5}
\hypersetup{colorlinks=true, citecolor=darkblue, linkcolor=darkblue, urlcolor=darkblue}

\usepackage{amsmath}

\title{State over Tokens:\\Characterizing the Role of Reasoning Tokens}

\author{
  Mosh Levy\\
  Bar-Ilan University\\
  \texttt{moshe0110@gmail.com}
  \And
  Zohar Elyoseph\\
  University of Haifa
  \And
  Shauli Ravfogel\\
  New York University \\ 
  \And
  Yoav Goldberg\\
  Bar-Ilan University \\ Allen Institute for AI
}

\begin{document}

\ifcolmsubmission
\linenumbers
\fi

\maketitle


\begin{abstract}
Large Language Models (LLMs) can generate reasoning tokens before their final answer to boost performance on complex tasks. While these sequences seem like human thought processes, empirical evidence reveals that they are not a faithful explanation of the model's actual reasoning process. To address this gap between appearance and function, we introduce the State over Tokens (SoT) conceptual framework. SoT reframes reasoning tokens not as a linguistic narrative, but as an externalized computational state—the sole persistent information carrier across the model's stateless generation cycles. This explains how the tokens can drive correct reasoning without being a faithful explanation when read as text and surfaces previously overlooked research questions on these tokens. We argue that to truly understand the process that LLMs do, research must move beyond reading the reasoning tokens as text and focus on decoding them as state.
\end{abstract}

\section{Introduction}

The assertion that Large Language Models (LLMs) can reason now appears unremarkable \citep{doi:10.1126/science.adw5211, maslej2025artificialintelligenceindexreport}. 
A key factor to achieve this was letting models generate a sequence of tokens before their final answer, which significantly improves performance \citep{wei2022chain,Zelikman2022STaRBR,deepseek2025r1}.

We refer to this sequence of symbols, which includes phrases such as `therefore', `consider' and `it follows that' as the \textbf{reasoning tokens}, and explicitly distinguish this name from \textbf{reasoning text}, which is the same tokens when interpreted by a reader according to their English semantics. 

The combination of (a) utility in improving the answer; and (b) appearance as a readable English text, may lead to the following inference:  the reasoning text is a faithful explanation of the model's reasoning process. This is strengthened by metaphors like ``Chain-of-Thought'', which imply that the steps in the text are "thoughts" that explain the process. Yet empirical findings contradict this inference (see Section \ref{sec:not-a-window}): the reasoning text is not a faithful explanation of the model's reasoning process. While those findings clarify what the reasoning tokens are not, they leave a conceptual vacuum as to what they \emph{are}. Our aim in this paper is to help fill that vacuum. Drawing on the idea that metaphors structure understanding and guide thinking \citep{lakoff1980metaphors}, we believe that adopting more apt descriptions and metaphors can steer researchers and practitioners toward more fruitful directions and surface a new set of questions that are less salient under the prevailing view of the reasoning text as an explanation.

To understand reasoning tokens, we must focus on the functional \emph{role} they play, rather than their \emph{appearance}, which empirical research has found to be deceiving. To this end, we advocate viewing them as representing \textbf{State over Tokens (SoT)}, which characterizes the reasoning tokens as a computational device that enables the persistence of a process across separate and stateless computation cycles.
We argue that in order to understand the role of the reasoning tokens, we should interpret this sequence of tokens not using their semantics when read as English text, but as the state carriers of a computational process.

\begin{tcolorbox}[title=The Whiteboard Analogy]
Consider a hypothetical scenario: you are placed in a room with a problem written on a whiteboard. Your task is to solve it, but under a peculiar constraint: every 10 seconds, your memory is completely wiped and resets to the same state as it was when you first entered the room. Within each interval, you can read what is on the board and add a single word. These rounds repeat until you finally write down the solution. 
\\
\\
How might you solve a problem under such constraints? You may write intermediate results on the board: numbers, conclusions, or partial computations—that you can use when you return after being "reset". You might perform several mental calculations before writing down just the result so the \emph{whiteboard may not capture every calculation} that you did within each cycle. Moreover, you may use an encoding scheme when writing on the board: abbreviations, symbols, or even apparent gibberish that will mean something \emph{specific to you} when you encounter it in the next cycle. All in all, \emph{an outside observer may interpret the whiteboard text incorrectly}.
\end{tcolorbox}

The whiteboard analogy mirrors the model's operation: the words are the reasoning tokens, you are the model, and the ten-second interval represents the model's limited capacity per cycle. Motivated by this intuition, we present the SoT framework (Section \ref{sec:theoretical-framework}) and use it to demonstrate two common misconceptions that underlay the belief that the text is a faithful explanation (Section \ref{sec:dispelling-illusions}) and to examine its novel ontological divergence (Section \ref{sec:reasoning-text-duality}).

The SoT framework naturally gives rise to new research questions (Section \ref{sec:implications}), challenging current interpretability methods to account for multiple cycles, and questioning whether language is a special computational medium that can theoretically act both as a state and as a faithful explanation. 

Finally, this perspective compels us to reconsider established metaphors. While "Chain-of-Thought" captures the sequential nature of generation, it misleadingly suggests the text contains the complete reasoning process; SoT clarifies that tokens are the state passed between computations, and do not necessarily express the computations themselves. Similarly, the "Scratchpad" metaphor obscures a crucial distinction: while humans use a scratchpad to \emph{supplement} internal memory, for LLMs, the tokens are the \emph{sole} carrier of state between cycles.

By establishing reasoning tokens as externalized computational state, the SoT framework characterizes how LLMs employ what seems to us as natural language as a computational medium, and provides a foundation for future work on decoding how models use language tokens to reason.

\section{The Subject of Inquiry: Reasoning Tokens}
\label{sec:subject}

The subject of our discussion is the sequence of tokens that LLMs emit \emph{before} their final answer, which we refer to as \textbf{reasoning tokens} (which we distinguish from \textbf{reasoning text} which is the English interpretation of these same tokens). 
This includes scenarios where LLMs were guided to generate reasoning tokens using examples, instructions like ``let's think step by step'' \citep{nye2021workscratchpadsintermediatecomputation,wei2022chain,kojima2022large} or that they were trained to do so regardless of the input \citep{Zelikman2022STaRBR,openai2024o1,muennighoff2025s1simpletesttimescaling,deepseek2025r1, comanici2025gemini}. 

However, LLMs operate in the same way for the generation of the final answer as they do for the generation of reasoning tokens. Why separate the two when aiming to characterize the role of the produced tokens?

The first reason is techno-sociological: reasoning tokens appear to offer a second product beyond the answer itself. When interpreted as English, the \textbf{reasoning text} reads like a description of reasoning work, encouraging readers to see not only what the model answers, but also how it seemingly arrived there, serving as an apparatus for building trust \citep{jacovi2021formalizing,Ferrario2022HowEC}. 

The second reason is functional: we can conceptually distinguish the two stages by their training objectives. The final answer tokens are generated to be interpreted by humans (or external systems) as the solution to the user's query. During training, the answer text is optimized to reflect the solution, and thus has an additional constraint: it cannot explicitly reflect a computation. In contrast, the reasoning tokens are not constrained in this way, and their purpose is to support producing the final answer. Indeed, their existence improves the accuracy of the final answer \citep{kojima2022large,Zelikman2022STaRBR,muennighoff2025s1simpletesttimescaling, comanici2025gemini}. This distinction is especially pronounced in models trained with methods that optimize only the final answer \citep{lambert2024tulu,deepseek2025r1}, where the reasoning tokens are not enforced to any specific constraints as long as they serve the final prediction.

\subsection{Empirical evidence: plausibility without faithfulness}
\label{sec:not-a-window}

The reasoning text often reads like a logical derivation. As \citet{perrier2025typedchainofthoughtcurryhowardframework} notes, it can correspond to an explanation according to some deductive proof system, representing a sequence of steps that leads to a conclusion. In that sense, it serves as a \emph{plausible} explanation: it is convincing to human readers and presents \emph{a} valid path to the solution. However, while the sequence of outputs may correspond to a valid proof, it does not necessarily correspond to \emph{the} actual computation performed by the model. In other words, it fails to reliably be a \emph{faithful} explanation \citep{jacovi2021}---one that accurately reflects the true causal process driving the model's generation.

\paragraph{Incompleteness.}
The reasoning text can be \emph{incomplete} and omit critical factors that affected the model's final answer \citep{turpin2023language,yee2024dissociation,Chua2025AreDR,lindsey2025biology, arcuschin2025chain,marioriyad2025unspokenhintsaccuracyacknowledgement}. A study conducted in controlled settings showed that LLMs can even seem to follow alignment goals while omitting the undesired topics from the reasoning text \citep{skaf2025largelanguagemodelslearn}.

\paragraph{Semantic mismatch between LLM and reader.}
The semantic content of the reasoning text can be \emph{meaningless} or misleading from a human perspective. Some studies have found that LLMs can disregard specific details in the text they generate \citep{lanham2023measuring,paul2024makingreasoningmattermeasuring,chen2025reasoningmodelsdontsay}. Other work has shown that LLMs can be trained to produce irrelevant reasoning text and still provide accurate final answers \citep{stechly2025semanticsunreasonableeffectivenessreasonless,bhambri2025interpretabletracesunexpectedoutcomes,zolkowski2025earlysignssteganographiccapabilities}. This disconnect is also apparent to human readers, who fail to identify causal relations in the generation process of the reasoning text \citep{levy2025humansperceivewrongnarratives}. 

The peril lies in how apparent rationality can increase trust without justifying it, fostering over-reliance on model outputs. When reasoning text mimics systematic deliberation, users in high-stakes scenarios may be misled into unwarranted confidence precisely because the text implies a rigorous process \citep{ehsan2024explainability,rosenbacke2024explainable}.  

\subsection{What can be said on reasoning tokens}

Following this growing body of evidence and the practical risks it highlights, some have already argued that we should not treat reasoning text as a faithful explanation of the model's reasoning process \citep{agarwal2024faithfulness,sarkar2024large, barez-chain-2025, kambhampati2025stopanthropomorphizingintermediatetokens}. 
However, we believe that we need to characterize the role of reasoning tokens in a positive way, beyond merely stating that they are not an explanation.

What \emph{can} be said about the reasoning tokens? The following three statements hold: 

\begin{itemize}
    \item The reasoning tokens are an important part of the process that lead to the final answer.

    \item The reasoning tokens can sometimes be read as a series of steps that logically lead to the answer.

    \item The reasoning tokens are not a faithful explanation of the computation that led to the final answer.
\end{itemize}

We would like a metaphor that account for all three of these points. In what follows, we provide the State over Tokens view, which we believe does that.

\section{Conceptual Framework: State over Tokens}
\label{sec:theoretical-framework}

Rather than explaining the reasoning tokens by how they are perceived by human readers (how they \emph{look}), we consider their functional role within the mechanism of generation (what they \emph{do}). We view reasoning tokens as an accumulated computational substrate---a medium through which the computation state is encoded in tokens. 

\subsection{The formal framework}

Building on the intuition from the whiteboard analogy in the introduction---and consistent with how LLMs are implemented---we view the reasoning tokens not as a text with human-readable semantics, but rather as the evolving state of the reasoning process. The token sequence functions as state for the model, though it can also be interpreted as natural language text.

This state is what allows the LLM to maintain a coherent process across multiple computation cycles, each bounded in capacity, to solve problems that require more computation than any single cycle can provide.

To capture this architecture precisely, we view the autoregressive generation of the LLM as a recursive application of a pure function $\mathcal{M}(\cdot)$ on a token sequence (technically the LLM call is not pure as it involves a random sampling of tokens, we can assume the random seed is part of the call to remedy this). 

Each computation of $\mathcal{M}$ has limited computational capacity. While the computation capacity of the Transformer \citep{vaswani2017attention} does increase slightly with each input token \citep{merrill2022saturated,pfau2024letsthinkdotdot}, this is a fixed increase that does not depend on the content of the tokens.

The function $\mathcal{M}(\cdot)$ is repeatedly applied to a sequence of inputs following a deterministic process.
The initial input to the function is the user's input, a sequence of tokens we denote as $S_0$ (commonly referred to as the \emph{context} in LLM literature; we use \emph{input} to emphasize its functional role as the argument to $\mathcal{M}$):
\begin{align*}
    S_0 &= \text{user input}
\end{align*}
At each subsequent cycle $k$, the function $\mathcal{M}$ takes the current sequence $S_k$ as its input and produces a new token as output. This token is appended to $S_k$ to form $S_{k+1}$, the input for the next call:
\begin{align*}
    S_{k+1} &= S_k \oplus \mathcal{M}(S_k)
\end{align*}
Here, $\oplus$ denotes the concatenation operation. This continues until the final answer is produced.
Note that each prefix of the $S_k$ is an input (a computation state) of one of the prior cycles.

\paragraph{The function of the tokens.} This formalization reveals three key observations about the functional role of the tokens:
\begin{itemize}
    \item \textbf{The tokens are the only persistent artifact.} The sequence $S_k$ is the sole carrier of information between cycles. Internal states of the LLM exist only within each cycle and do not persist to the next, and will be recreated from scratch, exactly the same, in each cycle (modern Transformers use a key-value (KV) cache to avoid this recomputation, but the KV cache does not carry information beyond what would be recomputed from the tokens). Nothing else persists, and the LLM must reconstruct any needed information from $S_k$ alone.
    \item \textbf{The tokens exclusively dictate future computation.} The only information that the next computation of $\mathcal{M}$ receives is $S_k$. Thus, $S_k$ fully dictates what computation $\mathcal{M}$ can perform next. 
    \item \textbf{Encoding and decoding are internal to the LLM.} The way that $S_k$ affects the next computation and the overall process that leads to the answer is dependent on $\mathcal{M}$.
\end{itemize}

\paragraph{What the tokens enable.} 
The tokens are what enable the mechanism of a process that is composed of multiple cycles. Each individual computation is limited—it can perform only as much work as the LLM's depth allows in a single cycle. However, by encoding results into $S_k$ and building upon them in subsequent cycles, the computational capacity of multiple computations of the LLM is utilized. The tokens accumulation transforms the process from being bounded to one cycle to utilize the computation done in multiple cycles. This conceptual view is supported by theoretical findings, which demonstrate that—at least for specific constructions—accumulating tokens formally increases computational capacity \citep{merrill2024expressive,li2024chain}.

Following those observations, we argue that an accurate and revealing name for this is \textbf{State over Tokens} (SoT). We will use this term for the remainder of the paper.

\subsubsection{Implications of the State over Tokens view}
Viewing the reasoning tokens as state means we can attribute to them the properties inherent to the concept of "state" in computation. This reframing has several immediate implications for how we understand the tokens:







\paragraph{A state is forward-looking.} 
A state enables future computation, not describing past computations. While it is created by past computation, it does not determine it: various computations can lead to the same state.
Unlike a log that records past cycles, state encodes what is needed to continue the process. 

\paragraph{A change in the state will lead to a different process.} Altering the state will change the trajectory of future computations.

\paragraph{States are discrete snapshots.} The tokens seem to describe a flowing narrative, a text. But the LLM does not process this flow—at each cycle, it operates on a single prefix as a discrete state. Reading continuously obscures how each prefix functions in the process, which may differ from what the tokens appear to mean as a narrative.

\paragraph{A state is necessarily partial.} As state, the tokens need only contain what is required for the next computation to proceed with the overall process. The bulk of each cycle's computation does not need to be externalized (see Section \ref{sec:misconception-of-completeness} for further discussion).

\paragraph{A state is created to be used by its creator.} What matters is not what a generated token means to human readers, but how the LLM decodes it uses it in subsequent cycles. The LLM can use its own semantics independent of natural language interpretation (see Section \ref{sec:misconception-of-shared-meaning}).

\section{Two Misconceptions Behind the Illusion of Explanation}
\label{sec:dispelling-illusions}


To examine the nature of the relation between a sequence of states and the computation that produced it, it is instructive to take a step back and consider recursive iterative processes where $\mathcal{M}$ is not an LLM but a simpler function.

As an example, consider a process that computes the Catalan numbers, a sequence of natural numbers occurring in various counting problems. The Nth Catalan number is defined recursively by the formula
$C_n = \sum_{i=0}^{n-1} C_i C_{n-1-i}$ with base case $C_0 = 1$.
The recursive definition can be translated into an iterative one, in which a pure function 
takes a prefix of the sequence and computes the next value according to the recursive definition.
When each number is computed, it is added to the state that becomes the input of the next function call. The function $\mathcal{M}$ takes as input a target index $N$ and a prefix of $N$ values (the first $N$ Catalan numbers), and returns the $(N+1)$-th Catalan number. The computation state in our case is the sequence of $N$ already computed numbers. Each computed value is concatenated to the input sequence; the resulting sequence becomes the state for the next call. 
For example, computing the 6th Catalan number generates the sequence of states:

\begin{align*}
    S_0 &= \text{6th?} & \\
    S_1 &= S_0 \oplus \mathcal{M}(S_0)  = S_0 \oplus \mathcal{M}(\text{6th?}) &=& \text{ 6th?} &\oplus& \text{ 1} \\
    S_2 &= S_1 \oplus \mathcal{M}(S_1) = S_1 \oplus \mathcal{M}(\text{6th?, 1}) &=& \text{ 6th?, 1}  &\oplus& \text{ 1} \\
    S_3 &= S_2 \oplus \mathcal{M}(S_2) = S_2 \oplus \mathcal{M}(\text{6th?, 1, 1}) &=& \text{ 6th?, 1, 1} &\oplus& \text{ 2} \\
    S_4 &= S_3 \oplus \mathcal{M}(S_3) = S_3 \oplus \mathcal{M}(\text{6th?, 1, 1, 2}) &=& \text{ 6th?, 1, 1, 2} &\oplus& \text{ 5} \\
    S_5 &= S_4 \oplus \mathcal{M}(S_4) = S_4 \oplus \mathcal{M}(\text{6th?, 1, 1, 2, 5}) &=& \text{ 6th?, 1, 1, 2, 5} &\oplus& \text{ 14} \\
    S_6 &= S_5 \oplus \mathcal{M}(S_5) = S_5 \oplus \mathcal{M}(\text{6th?, 1, 1,  2, 5, 14}) &=& \text{ 6th?, 1, 1, 2, 5, 14} &\oplus& \text{ 42} 
\end{align*}

Note that while the numbers 1,1,2,5 and 14 are all \emph{needed for} computing 42, and were \emph{used in the process of} computing 42, we intuitively \emph{do not consider them} as "an explanation for how 42 was computed". Indeed it actually seems odd for us to treat them as an explanation. But why would it be different for the intermediate steps of an LLM? This form of alienation is already instructive. But let's now consider two concrete implications.

\subsection{The misconception of completeness}
\label{sec:misconception-of-completeness}

It should be clear to any reader that while the
intermediate sequence $1, 1, 2, 5, 14$ is crucial for computing $42$, this sequence results from the computation and allows it to continue, but it \emph{is not} the computation itself. The sequence of values does not equal the computation that took place for creating them. Additionally, this sequence does not determine a computation: we cannot infer the computation from the sequence (we can guess but cannot be sure), and it is clear that the computation steps that were involved in computing each number are not reflected in the output sequence. We see results of intermediate computations, not the algorithm that is used for producing them.

We can also imagine a function $\mathcal{M}'$ that is only called 3 times and not 6, with the intermediate values 1,2,5. This function computes two following numbers at a time, and only outputs the second one. In each invocation, it recomputes the missing numbers, and continues the sequence. Even if we do choose to treat the steps as an explanation, there is no guarantee that this explanation is complete.

The SoT functions as scaffolding that propels the process forward; a misconception arises when we mistake this scaffolding for the building.

\subsection{The misconception of shared meaning}
\label{sec:misconception-of-shared-meaning}
Even if we accept that the state shows only partial results, we might assume the LLM interprets them as humans do. However, this assumption is also mistaken. 

Consider an alternative function for computing the N-th Catalan number, whose final state is not $1, 1, 2, 5, 14, 42$ but rather $11, 11, 12, 15, 24, 52$. Such a function could be just as effective: at each step, $\mathcal{M}$ will subtract 10 from each of its inputs, perform the computation, add 10 to the result, and return it as output. At the final (N-th) stage, it will subtract 10 from the last item and report it as the answer.

This simple additive transformation demonstrates that the reasoning tokens in the SoT---the very symbols we read and interpret---may function in a way entirely different from what human readers naturally ascribe to them. Even elementary encoding schemes can render the surface meaning of text opaque to human interpretation while remaining fully usable for the LLM.

\paragraph{Beyond numerical encoding.} This encoding arbitrariness extends beyond numerical computation to the full range of SoT. When an LLM writes "I need to reconsider this approach", the phrase might function not as genuine meta-cognitive reflection, but as an encoding for something else. Given empirical findings on how models can embed functionally relevant information in ways that remain opaque to human readers \citep{cloud2025subliminallearninglanguagemodels}, it is likely that real-world encodings are far more intricate than a simple additive shift. This view, together with Section \ref{sec:misconception-of-completeness}, may also give intuition to the phenomenon where LLMs succeed at evading mentioning relevant topics in their reasoning text \citep{emmons2025chainthoughtnecessarylanguage, li2025llms}.

\section{The Ontological Divergence: Text vs. State}
\label{sec:reasoning-text-duality}

We argue that the divergence between form and function in reasoning LLMs---that the same tokens can be interpreted (by humans) as English text while at the same time also interpreted (by the LLM) as an encoding of a computation state for an iterative function---is a novel, and hence \emph{foreign} phenomenon in human experience. While we are accustomed to symbols that can be interpreted in multiple ways (the word "apple" can refer to either a fruit or a company, the same sequence of bytes can be interpreted as either a floating point number or a memory address), the State over Tokens case is fundamentally different.

The reasoning tokens go beyond admitting multiple interpretations; they inhabit distinct ontological categories. To a human reader, the sequence is a text, parsed according to linguistic conventions to extract semantic meaning. To the model, however, the same tokens serve as a computational substrate, accumulated one token at a time to mechanically drive the next step of a process. These are not different perspectives on the same underlying content, but completely different kinds of entities—a communicative medium and a functional state—that happen to share the same embodiment. We believe this is how the tokens can function as a computational substrate without becoming \emph{about} the computation in a way that human readers can discern by reading. The LLM uses a sequence of tokens that can be interpreted by humans according to one semantic system, while at the same time it is interpreted by the LLM using an alternative semantic system.

\section{Questions Raised by State over Tokens}
\label{sec:implications}

\subsection{What SoT means for interpretability}

Rather than solving the challenges that existing interpretability work addresses---such as understanding model internals, tracing how features emerge, or identifying mechanisms \citep{alain2016understanding,li2023emergent,bereska2024mechanistic}---the SoT view surfaces a distinct, new challenge: investigating how LLMs construct and use computational state through tokens.
We call for \emph{recognizing the encoded state as a first-class object of study}. This approach neither reads the reasoning tokens as ordinary English text nor dismisses them as an inadequate explanation, but rather acknowledges that understanding how LLMs use state requires work to decode the structure and function of the token sequence. The new overarching question becomes "How is the computation state encoded in the tokens being maintained and used by the LLM?" This shift in perspective opens new questions about the structure and behavior of the encoded state itself: 
\begin{itemize}
    \item How do LLMs decide what information to externalize at each cycle?
    \item What information is actually encoded in the state at different points in the sequence?
    \item Do they use consistent encodings across the solution of a problem and across different problems?
    \item How does information propagate through the sequence?
\end{itemize}
Understanding these properties requires decoding the relationship between internal computation and the externalized token-based state.

Recent interpretability work provides a starting point—for instance, studying which parts of the sequence are most critical for the final answer \citep{bogdan2025thoughtanchorsllmreasoning} or analyzing model components' activity during different cycles  \citep{dutta2024thinkstepbystepmechanisticunderstanding,chen2025doeschainthoughtthink,zhao2025verifyingchainofthoughtreasoningcomputational}.
However, the grand challenge of understanding how LLMs manage state through tokens remains largely open.

\subsection{Is language special for SoT?}

The SoT framework also raises the question of the \emph{arbitrariness of the medium for discrete, token-based computation}. We previously argued that SoT can, in principle, encode arbitrary computation regardless of its surface semantics. But to what extent is this actually the case? Are all media equally expressive, or is using natural language sequences particularly well-suited for encoding the kinds of computations that LLMs perform?

One can hypothesize that the massive pretraining stage on natural language data induces an inductive bias to reason \citep{venhoff2025basemodelsknowreason} in ways that are \emph{in accordance with its semantics}, which makes the reasoning text read as a plausible explanation. Under this view, elaborate encoding schemes may either contradict the training distribution---where SoT updates often gradually lead to the correct continuation---or require additional computation that is disfavored given the pretraining data. If this hypothesis is correct, LLMs possess at least an inductive bias to represent state in a manner that reflects their internal computation. 

In this context, recent works have explored alternative media for state representation, ranging from simple vectors \citep{hao2025traininglargelanguagemodels,butt2025softtokenshardtruths,hwang2025latentreasoningsentenceembedding} to structured formats \citep{domingos2025tensorlogiclanguageai}. However, the question of whether natural language offers unique advantages over these alternatives remains open. 

\subsection{Can SoT ever be a faithful explanation?}

Finally, the SoT view sharpens the question of whether the faithfulness of reasoning tokens can be improved. If we aim to make the text a faithful explanation, there will be a fundamental tension: the reasoning tokens will have to serve two distinct masters. Functionally, as a state, the tokens are optimized to encode the information necessary to drive the next computation cycles. Interpretively, as an explanation, we want them to also describe the computation itself---whether they describe the computation that has already occurred in past cycles, or the computation that will occur in future cycles.

This dual requirement creates a bottleneck. The tokens are the only medium available for both the computation \emph{and} its description. Requiring an LLM to explain its own computation while simultaneously performing it amounts to a form of \emph{metacognitive} capability; the LLM must reason about its own reasoning process within the same sequence that sustains it. For the explanation to be faithful, the information the model encodes to solve the problem (the state) must be identical to the information a human extracts by reading the text (the explanation). However, an optimal computational state might require encoding information that is redundant, non-linear, or semantically opaque to a human reader. Conversely, constraining the state to be a coherent, linear English narrative might strip it of the information needed for the reasoning process.

Thus, the challenge is not merely whether LLMs can be trained to produce plausible explanations, but whether the medium of natural language tokens has the capacity to simultaneously function as an efficient computational substrate and a transparent descriptive record. This forces us to confront a fundamental question: can the same sequence of symbols simultaneously carry the full weight of a machine's computation and transparently reveal the logic of that process to humans?

The question extends also to viewing reasoning texts as \emph{rationalizations}: texts that consist of a sequence of steps leading to the answer, which can be used by a human reader to verify it. To what extent can a sequence be used concurrently both as carrying state and as providing an effective rationalization?

\section{Conclusion}
\label{sec:conclusion}

The reasoning tokens generated by LLMs before an answer are best understood not as a narrative of thought, but as an externalized computational state, a role captured by the term \textbf{State over Tokens (SoT)}. 

Adopting this perspective resolves the persistent misconceptions that underlie the illusion of explanation. The \emph{Misconception of Completeness} is explained by recognizing that SoT externalizes only what is functionally necessary for the next cycle, not a full account of the computation that occurred. The \emph{Misconception of Shared Meaning} is exposed by the insight that the text's human-readable meaning can be incidental to its computational function, which may rely on arbitrary, model-specific encodings. This semantic mismatch—where the token sequence functions as computational substrate while appearing as natural language—represents an unprecedented ontological divergence: two fundamentally incompatible modes of interpretation coexisting within the same textual artifact.

While much research has focused on what this text is not, the SoT view attempts to explain what it \emph{is}. This opens concrete research directions—understanding how LLMs construct and use state through tokens, investigating whether natural language possesses unique advantages as a computational medium, and exploring whether SoT can ever faithfully explain the underlying computation. Recognizing the reasoning tokens as a computational state provides a clear basis for calibrating trust and for investigating the new questions this view opens.

\bibliography{colm2025_conference}
\bibliographystyle{colm2025_conference}

\end{document}